\documentclass[eng]{FCEFyN-class}
\usepackage{hyperref}
\hypersetup{ 
    urlcolor=blue
    }
\title{\LARGE{Winning Through Simplicity: Autonomous Car Design for Formula Student}}

\shorttitle{Autonomous System Elefant Racing e.V. 2021}
       
\author[1]{Tobias Friedrich}
\author[1]{Marco Müller}
\author[1]{Adrian Bauske}
\author[1]{Simon Härtl}
\author[1]{Johannes Herrmann}
\author[1]{David Förster}
\author[1]{Tobias Tietze}
\author[1,2]{Sebastian Sartor}

\usepackage{booktabs}
\usepackage[english]{babel}
\usepackage{csquotes}

\affil[1]{Elefant Racing e.V. at University of Bayreuth, Bayreuth, Germany}
\affil[2]{sebastian.sartor97@gmail.com; +49 174 68694944}

\firstauthor{Friedrich}

\thismonth{June}
\thisyear{2024}

\usepackage{tabularx}

\newenvironment{conditions*}
  {\par\vspace{\abovedisplayskip}\noindent
   \tabularx{\columnwidth}{>{$}l<{$} @{${}={}$} >{\raggedright\arraybackslash}X}}
  {\endtabularx\par\vspace{\belowdisplayskip}}

\usepackage{biblatex}
\begin{document}
\small
\abstract{This paper presents the design of an autonomous race car that is self-designed, self-developed, and self-built by the Elefant Racing team at the University of Bayreuth. The system is created to compete in the Formula Student Driverless competition. Its primary focus is on the Acceleration track, a straight 75-meter-long course, and the Skidpad track, which comprises two circles forming an eight. Additionally, it is experimentally capable of competing in the Autocross and Trackdrive events, which feature tracks with previously unknown straights and curves. The paper details the hardware, software and sensor setup employed during the 2020/2021 season. Despite being developed by a small team with limited computer science expertise, the design won the Formula Student East Engineering Design award. Emphasizing simplicity and efficiency, the team employed streamlined techniques to achieve their success.}
%
%
\maketitle
\thispagestyle{fancy}
%

\section{Introduction}
Building on the pioneering efforts of other Formula Student teams, such as those documented in \autocite{AMZFullyAutonomousRacecar}, \autocite{valls2018design}, \autocite{Zeilinger2017}, and \cite{8462829}, we have developed an autonomous system for the race car of the Elefant Racing team at the University of Bayreuth. The first section outlines the constraints and principles that guided our design process. Subsequent sections detail the derived hardware setup and software modules.
\subsection{Design Constraints and Objectives}
The autonomous system design follows the following constraints (C) and objectives (O):
\begin{description}
\setlength\itemsep{-0.3em}
\item[(C1) First-year team] As a first-year FSD team, there is no internal knowledge to build on. \label{C1}
\item[(C2) Small team] There is a smaller number of students ($40-60\%$ of large teams), mainly pursuing the Bachelor's degree.  
\item[(C3) Few software developers] There are only 3 to 4 active students with a software developing background ($10-30\%$ of large teams, e.g. \href{https://driverless.amzracing.ch/en/team/2021}{AMZ} or \href{https://tufast-racingteam.de/teams/team-2021/}{TUFast})
\item[(C4) Single old car for FSD and FSE with little space] The team also participates in Formula Student Electric (FSE) competitions without an autonomous system. This is done with the same car due to (C2). Hence, the autonomous system must be easily attachable and detachable. It must not affect the performance when removed. The two-season building phase of the base vehicle due to COVID-19 means that it has not been designed with driverless components in mind.
\item[(C5) No aerodynamic, weight, or power concerns] The targeted low speed and acceleration (40-50 km/h in Acceleration and 15-20 km/h in Skidpad) for the first driverless vehicle make these not necessary. A powerful 1000W DCDC provides a reasonable power budget.
\item[(O6) Safety \& Rule Compliance] Only a quarter of all qualified teams competed in at least one dynamic discipline at Formula Student Germany in 2019 \cite{FSGResults2019}. This was mainly due to hardware failures or rule non-compliance due to an insufficient safety concept. 
\item[(O7) Robust against sensor errors] A reliable solution has to deal with sensor errors in perception and state estimation.
\item[(O8) Simplicity] Considering (C1, C2, O6) the simplest solution with the lowest complexity should be chosen: "keep it simple, stupid (KISS)". It ensures easy knowledge transfer, easy integration of new members, and a short development cycle.
\item[(O9) Extensible] When possible and not contradicting (O8), methods and hardware should be chosen that could be reused, extended and improved in the future.
\item[(O10) Good direct sensor data instead of algorithms] In view of (C2, O7), the hardware should be carefully chosen such that the algorithms are as easy as possible (O8) and provide a good foundation for future systems (O9). In particular, relevant data should be directly measured instead of estimated when possible.
\item[(O11) Designed for Acceleration and Skidpad] Considering (C2, O8), the vehicle must compete in these two easier disciplines in Formula Student. Participation in Autocross and Trackdrive is experimental.
\end{description}

\section{Autonomous Hardware}
\begin{figure}[h]
\includegraphics[width=0.5\textwidth]{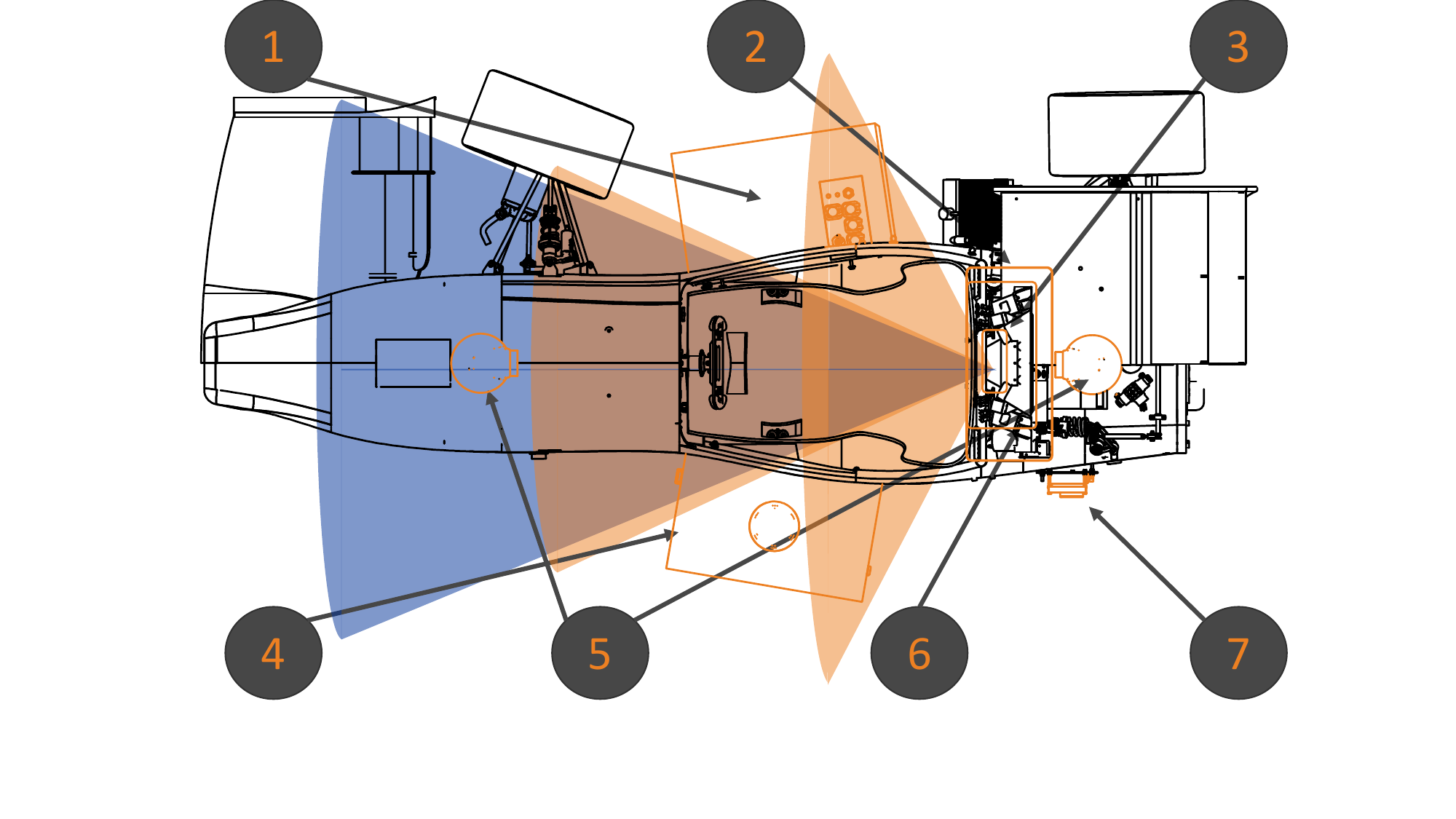}
\caption*{
1: Driverless Compute Unit (DCU); 2: Emergency Brake System (EBS); 3: GNSS/IMU-system; 4: Electronic Box with LTE router; 5: GNSS antennas; 6: Monocular cameras (near: 125° FOV; far: 50° FOV) in orange and stereo camera (45° FOV) in blue; 7: Ground Speed Sensor}
\caption{Autonomous hardware}
\label{fig:hardware_overview}
\end{figure}
The autonomous hardware is shown in Figure \ref{fig:hardware_overview}. The boxes at the side replace delicate aerodynamic components that might get damaged when hit by cones. One rear aerodynamic wing is replaced by a carbon plate with a larger GNSS antenna and a cutout for the camera system. All components are easily attachable by connecting a few plugs and fastening some bolts (C4, C5).
\subsection{Electric Motors for Acceleration and Deceleration}
The vehicle is equipped with four electric wheel hub motors that are controlled by the torque vectoring system. It is capable of braking by recuperation. In particular, during normal autonomous operation the car is decelerated only with the motors without the hydraulic disk brakes.
\subsection{Emergency Brake System (EBS)}
\begin{figure}[h]
\includegraphics[width=0.45\textwidth]{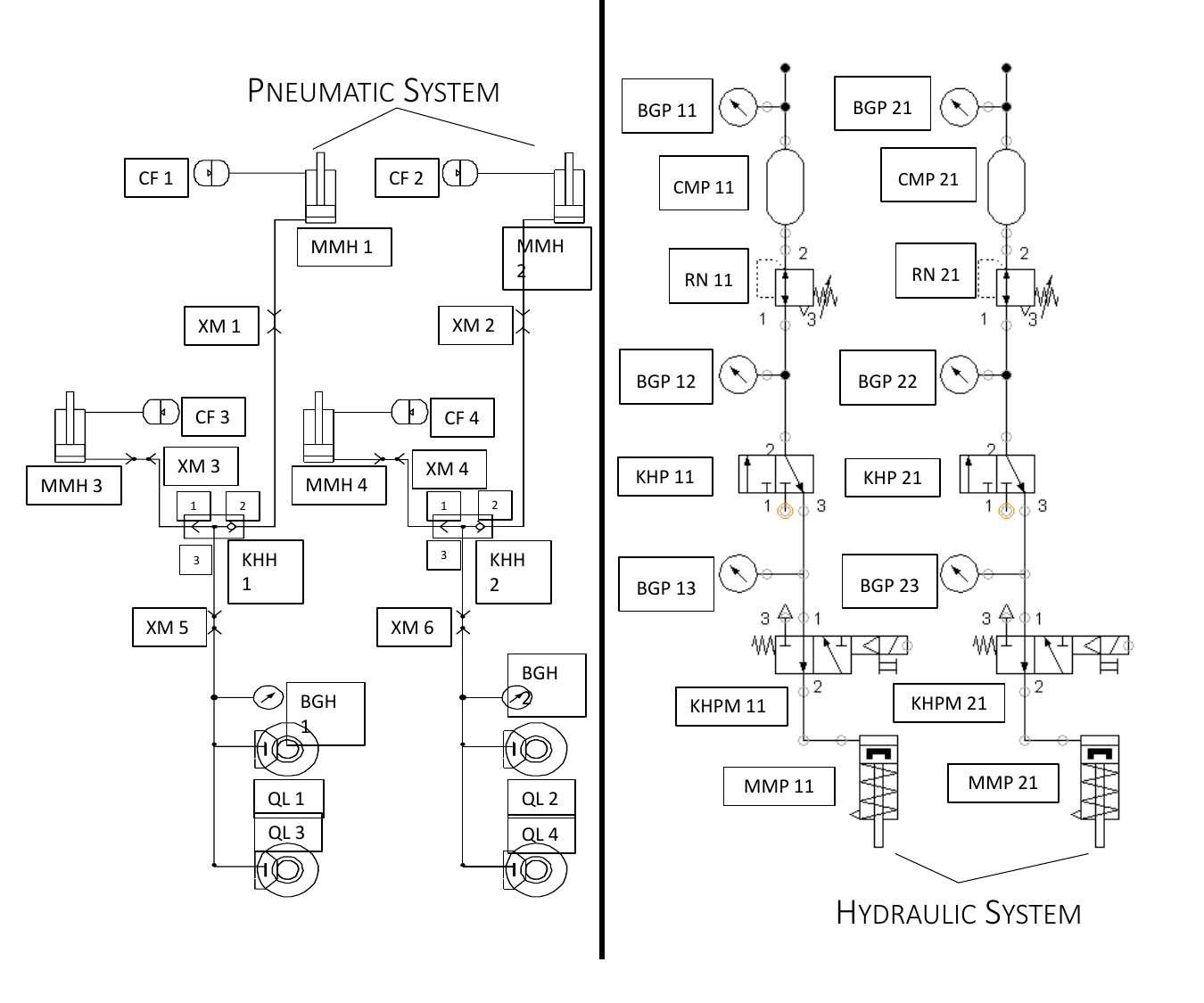}
\caption{Pneumatic and Hydraulic Circuits of EBS}
\label{fig:EBS}
\centering
\end{figure}

In view of (O6), the EBS safety concept has received much attention: The emergency brake system is comprised of two completely independent pneumatic and hydraulic circuits for the front and rear brake circuit, respectively (see Figure \ref{fig:EBS}). In case of an emergency, pressurized air from tanks flows into pneumatic cylinders. They independently press on two brake master cylinders (MMP11, MMP21) for the EBS. An EBS brake master cylinder (MMH1 or MMH2) and a brake pedal master cylinder (MMH3 or MMH4) are connected to the inputs of a shuttle valve. The higher of both pressures is available at the output which is connected to the brakes. \\
A check-up-sequence runs before starting the car in manual mode or autonomous mode. It checks that the EBS tanks are without pressure or that all parts of the EBS are ready and pressurized, respectively. Considering (O6), the software is tested against a simple dynamic simulation of the pneumatic circuit. To this end, the system has been modelled by an equivalent electric circuit made out of resistors, capacitors, and two-ports and integrated with a simple forward Euler-scheme.\\
The sequence finds all modelled single component failures and most failures with at most three failing components (see Table \ref{tab:EBS_failures}). Testing all failure combinations was infeasible due to combinatorial explosion. It was found that the pressure tank has to be supervised by at least two pressure sensors to guarantee that the tanks are pressure-free for manual mode. Ideally, both sensors should be directly connected to the tank. However, the rules mandate a pressure regulator directly on the tank.
\begin{table}[h]
    \centering
    \begin{tabular}[t]{c|c|c}
 \toprule
 Connection &   \shortstack{Pressure \\ Regulator} & Electric Valve  \\ 
  \midrule
 large leakage & no regulation & always open \\ 
 small leakage & too high & always closed \\ 
 partially blocked & too low & no reset to open\\
 blocked & low flow-rate & \\
 \bottomrule
 \toprule
 Manual Valve & Pressure Sensor & \shortstack{Air Cylinder \\ w/ Hydraulics} \\
 \midrule
 wrong position & output disconn. & hydr. leakage \\
& \shortstack{constant wrong \\ output \\ (different levels)} & \shortstack{wrong transfer \\ function}\\
 \bottomrule
    \end{tabular}
    \caption{EBS failures for unit testing}
    \label{tab:EBS_failures}
\end{table}

\subsection{Steering Actuator}
\begin{figure}[h]
\includegraphics[width=0.45\textwidth]{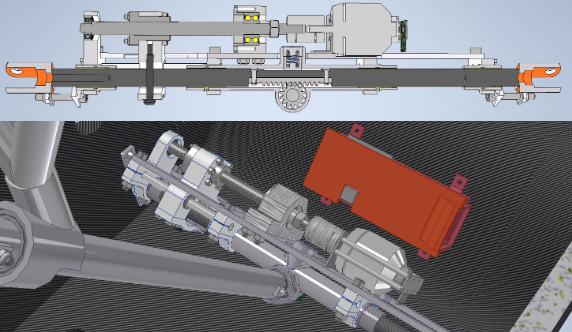}
\caption{Steering actuator with cross section and placement}
\label{fig:steering}
\centering
\end{figure}

The steering actuator is located at the steering rack. It consists of a rack parallel ball screw driven by a brushless servo motor. It is coupled to the steering rack via a cross bore and a shoulder screw. The ball screw's low coefficient of friction and lack of self-retention enables manual steering in manual mode. The assembly can easily be removed from the car by removing the eight bolts and the shoulder screw (Figure \ref{fig:steering}).
This actuator is able to steer from fully left to fully right in 0.4 seconds. Providing 2000 N at the rack to steer while the car is not moving. Compared to past telemetry data from previous Autocross runs, this reaches the performance of a manual driver 99 percent of the time.
\subsection{Processing Units \& further Electronics}
All additional electronic processing components are located within two ventilated boxes at the side of the car. This includes the Driverless Compute Unit (DCU) made out of consumer miniature PC components (AMD Ryzen 3700X, Nvidia 2070 Super). The boxes are designed larger than needed to plan for extension or replacement (O9). The GNSS system profits from RTK correction data from the internet, provided by a dual-SIM cellular router. As a fallback, it can route the traffic over the long range directional antenna to an off-track stationary cellular router. This connection is also used for telemetry data.
\subsection{Sensors}
\subsubsection{GNSS, IMU and ground speed velocity sensor}
The state of the vehicle, in particular the (vehicle-frame) velocity, has to be estimated (see State Estimation). The velocity is measured with a dual-antenna GNSS system (Novatel PwrPak7D-E2) in case of good GNSS reception. To deal with the likely case that the GNSS signal fails, it is also precisely measured with an optical ground speed sensor (Kistler Correvit SFII) (O7, O10). This allows redundancy without a complex vehicle model with unknown parameters, without using wheel sensors and without estimating tire slip (O8, O10). The inertial measurement unit (IMU) is needed to observe the vehicle state fully. It is integrated into the GNSS receiver, which automatically calibrates and corrects the IMU (O8, O10). In the case of good GNSS reception, accurate absolute position and heading measurements from the dual-antenna-RTK-GNSS support localization, testing, and data collection.
\subsubsection{Camera and LiDAR system}
For cones nearby, two slightly overlapping cameras (Basler acA1920-40 with a Sony IMX249 CMOS 2.3 MP sensor) with combined HFOV of $125 ^{\circ}$ are used. This has the downside that some cones are seen by both cameras. So matching and deduplication is needed. For easy matching the parallax effect between both cameras is minimized by placing the entry pupils of both cameras as near as possible to each other. \\
Alternatively, a smaller HFOV ($100-110 ^{\circ}$) and a smaller effective resolution per cone could also be reached with a single camera and an extreme wide-angle lens. This would have the benefit of lower complexity (O8) since each cone is only captured once. However, during corners, most curve-inside cones are near the edge of the image where the distortion is the largest in particular for a wide angled lens. Moreover, this provides no room for improvements (O9), so we refrain from this solution.\\
Depth estimation (with errors of $~0.3-3m$) is rather challenging for cones further away (10-20m) with a single camera. So additional sensors are needed. \\
LiDAR systems provide precise depth measurements almost irrespective of distance and lighting conditions. Most LiDARs scan the points consecutively and with a lower (vertical) resolution. This requires ego-motion-correction with a dependency on velocity and yaw rate. Moreover, precise time synchronization is needed. This makes early sensor fusion with reprojection of LiDAR data into camera images complex, especially for the considerable small cones. In late sensor fusion, the cones are detected independently in LiDAR data and camera images. This requires an additional pure LiDAR detection pipeline and a fusion step. Due to (C3, O8) we refrain from this solution.\\
However, flash LiDARs acquire all points in a single shot (cf. global shutter) and with a higher vertical resolution. This makes reprojecting the points into camera images easy when both sensors capture at the same time. This is an ideal solution since depth data can just be considered an additional channel in the image (C3, O8, O10). \\ 
However, some Flash-LiDARs are rather large and did not fit into the car without heavy modification (C4). Others were not available or too expensive, especially when excepting new models and price reductions in the future. With this in mind, we assume that next season a flash LiDAR can be used. \\
As a drop-in replacement, this season, the stereo camera Nerian Karmin3D with the stereo matching SceneScan Pro is used (O10) for cones further away. Despite challenges mounting the larger camera rule compliant to the car, a larger base length of 25cm has been chosen for more precise depth estimates ($0.4m$ at $20m$) (O10). It is accompanied by another mono-ocular camera for future experiments, when the camera is replaced by a LiDAR.
\subsubsection{Synchronization and Calibration}
The processing units, including the Nerian SceneScan Pro, are synchronized over Precision Time Protocol (PTP). The Nerian SceneScan Pro triggers all cameras at the same time and sends the timestamped depth and color images over Ethernet to the DCU, where they are matched with the image data from the USB cameras. The ground speed sensor data is timestamped when received over CAN.\\
The cameras are extrinsically calibrated to each other using the usual checkerboard pattern. The heading of the cameras with respect to the car axis is done with a checkerboard placed exactly on the vehicle approximately $10\textit{m}$ away. The remaining position and orientation of sensors and antennas relative to each other and the vehicle are directly measured using a total station and digital inclinometers. This makes the development of calibration software unnecessary (C2, O6).

\section{Testing \& Data collection}
\begin{figure}[h]
\includegraphics[width=0.45\textwidth]{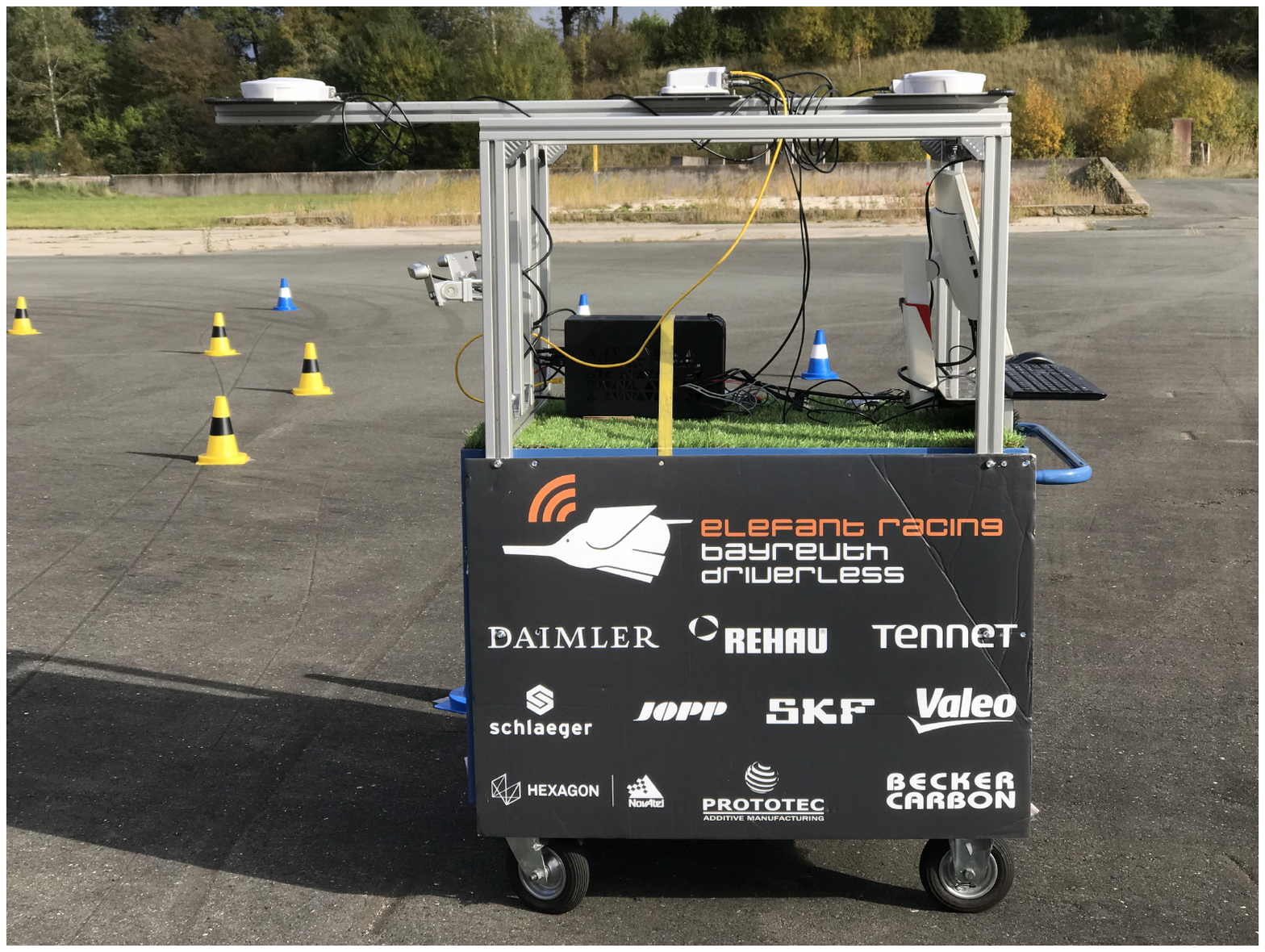}
\caption{Test Platform}
\label{fig: test_platform}
\end{figure}
Parts of the software have been tested on a test platform (see Figure \ref{fig: test_platform}) equipped with all the sensors, batteries, and the compute units. Moreover, this setup is used to collect images to train the neural network of the Perception module. Using another RTK GNSS rover allows to set up official tracks faster. Moreover, the global position of each cone is saved centimeter-accurate. The position and heading of the test platform is also measured by an RTK GNSS system. This allows validating and testing of the Localization algorithm and the depth estimation in the perception module. Additionally, images of cones in various environments have been taken.
\section{Software}
%
%
%
%
\subsection{State Estimation}
\label{sec:state_estimation}
The state of the vehicle is estimated with an Extended Kalman Filter (EKF). The main consumer of the state is the Localization and Mapping module, which uses \footnote{frames and variable names (almost) according to ISO 8855}:
 \vspace{-2mm}
\begin{itemize}
\setlength\itemsep{-0.3em}
    \item velocity $\vec{v}$ in the vehicle-frame (or the change of position in local vehicle coordinates), in particular the longitudinal velocity $v_x$
    \item yaw rate $r=\dot{\psi}$  (or the change of the heading $\psi$),
    \item optionally if available the position $\vec{P}=(X,Y)$ and heading $\psi$ w.r.t.\ an earth fixed ENU coordinate system
\end{itemize}
 \vspace{-2mm}
\noindent
The Kalman Filter has to be designed such that it provides locally correct estimates for $\vec{v}$ and $r$ with either the GNSS/IMU-system or the ground speed sensor and the IMU only. The GNSS should update estimates for earth fixed quantities $\vec{P}$ and $\psi$ whenever available. \\
To this end, we define state $\vec{x}_{EKF}=(X,Y,\psi,v_x,v_y,r,a_x,a_y)^t$, where $\vec{a}=(a_x, a_y)$ is the acceleration in the vehicle-frame. For the predict step, the vehicle is modelled as a rigid body with almost constant (earth fixed) acceleration and almost constant yaw rate. After a partial coordinate transform, this is written as
\begin{align*}
    \dot{X} &= \cos(\psi) v_x - \sin(\psi) v_y + \varepsilon_X \\
    \dot{Y} &= \sin(\psi) v_x + \cos(\psi) v_y + \varepsilon_Y \\
    \dot{\vec{v}} & = \vec{a} + r (v_y, -v_x)^t + \varepsilon_v\\
    \dot{\vec{a}} & = \varepsilon_a\\
    \dot{\psi} & = r + \varepsilon_\psi\\
    \dot{r} & = \varepsilon_r
\end{align*}
with independent process noise $\varepsilon_X, \dots, \varepsilon_r$.
Experimentally, a kinematic bicycle model \cite{KongPfeifer2015} can be used instead (see below).\\
The measurement model is formulated with sensor data already corrected for misalignment and the GNSS/IMU-receiver (transformed into) in the vehicle origin. The vehicle-frame position of the ground speed sensor (GSS) is $\vec{p}_{GSS}$.
\begin{align}
    \label{eq:GNSS_pos}
    (X_{GNSS}, Y_{GNSS})^t &=  (X, Y)^t + \varepsilon_{P, GNSS} \\
    \label{eq:GNSS_heading}
    \psi_{GNSS} &=  \psi + \varepsilon_{\psi, GNSS} \\
    \label{eq:GNSS_vel}
    \vec{v}_{GNSS} &=  \vec{v} + \varepsilon_{v, GNSS} \\
    \label{eq:IMU_a}
    \vec{a}_{IMU} &=  \vec{a} + \varepsilon_{a, IMU} \\
    \label{eq:IMU_r}
    \vec{r}_{IMU} &=  r + \varepsilon_{r, IMU} \\
    \label{eq:GSS}
    \vec{v}_{GSS} &=  \vec{v} + r \vec{p}_{GSS} + \varepsilon_{v, GSS}
\end{align}
Note that the IMU and GNSS measurements are already fused by the receiver. For simplicity, we nevertheless assume independent measurement noise. In the best case, all sensors work properly and all measurements \eqref{eq:GNSS_pos} to \eqref{eq:GSS} are integrated. If the GNSS sensor fails, e.g. due to signal loss, only \eqref{eq:IMU_a} to \eqref{eq:GSS} are used for the update step. The important state variables  $\vec{v}$ and $r$ remain observable. The earth fixed state variable begins to drift. If only the GSS fails, e.g. due to water on the track, we rely on the GNSS/IMU-system with a minor loss in accuracy mainly for $v_y$. If the GNSS and the GSS both fail, we switch from the rigid body model to the kinematic bicycle model using only the IMU sensor.
%
%
%
%
\subsection{Perception}
From the point cloud of the stereo-camera the ground is extracted using RANSAC. Then, the pose of the stereo camera relative to the ground is computed. Knowing the relative poses of all cameras from calibration this yields the pose of all cameras relative to the street. The other steps of the pipeline run separately on the data from all cameras in parallel.\\
First, the neural network CENTERNET \cite{duan2019centernet} is run on the color image. It extracts bounding boxes and key points for all cones in a single shot. The key points are extracted for easily distinguishable points (top of cone, corners of strip of cone).\\
In the stereo-camera, the depth data within the bounding box is clustered. The depth average of the cluster containing the midpoint of the bounding box, yields a depth estimate $d$ for a cone.
Since each cone's shape is known, the position $p_i$ of the point corresponding to a key point is known in a cone-local frame.
The position of a cone is determined by finding the position $(x,y,z)$ of the cone such that the sum of the following (weighted) errors is minimized
\begin{itemize}
\item Reprojection error for the key points of $p_i$ weighted with the variance matrix. The variance matrix is estimated from the errors between annotated key points and the neural network's output.
\item $(x-d)^2$ weighted with the factor $\lambda_1/x^2$
\item $z^2$ weighted with the factor $\lambda_2/(x^2+y^2)$ to force cones on the ground
\end{itemize}
%
%
%
%
\subsection{Localization and Mapping}
A simple and yet effective algorithm for mapping and localization is the FastSLAM algorithm which is easier to understand than other common SLAM algorithms (O8). Graph-based SLAM relies on a well-functioning front-end and an effective and robust backend which is hard to tune. ORB-SLAM consists of multiple, complex components and is available as open-source software\footnote{\url{https://github.com/UZ-SLAMLab/ORB_SLAM3}} however adapting it to Formula Student tracks is complicated.

FastSLAM version 2 uses visual information in the prediction step making it more precise however this only works if the visual information is more accurate than inertial information which is not the case for us\cite[p.~14]{FastSLAM} therefore we use version 1.

FastSLAM uses a particle filter where each particle has got information about its 2D pose and its map (consisting of cones)\cite[p.~9]{FastSLAM}. Each cone is represented by a 2D EKF\cite{enwiki:1004599218} and stores how often it has been seen ($n_s$) and has not been observed even though it should have been ($n_n$). The quality of a cone is determined by $\frac{n_s}{n_s + n_n}$ and deletes the cone if below 0.5. Colors detected by perception are summed up which gives sufficient information for the track detection.

Observed cones are matched to already existing cones by calculating the Mahalanobis function (see \ref{eq:mahalanobis-weight}) and randomly choosing one of the most likely associations. Only cones in a certain radius around the car are considered for matching to reduce computational complexity. As the car gets closer to the formerly ambiguous cones the weighting of particles automatically eliminates particles that made wrong data associations. Updating the cone position and covariance is done using the standard EKF way and incorporating the observation covariance in each step.

Resampling is done by using low variance resampling \cite[~p.3-4]{LowVarianceResampling} which is applied if the number of effective particles $n_{eff}$ drops below 50\% of the particle set size. Additionally, about 20\% of particles are chosen and their respective poses are randomly distributed to ensure particle exploration.

There are several weights (inspired by \cite{DBLP:journals/corr/abs-1809-10099}) which are used for determining the particle weight. If a cone is \textbf{n}ot \textbf{o}bserved even though in sensor range the weight  \textbf{$w_{no}$} is applied, if a cone is observed but \textbf{o}utside of sensor \textbf{r}ange the weight $w_{or}$ is used. For \textbf{n}ew \textbf{c}one insertions $w_{nc}$ is taken (different for each discipline). \textbf{M}atched cones are given the weight $w_m$ ($z$ is the observed cone position, $\hat{z}$ the previous cone position) with covariance $\Sigma$ which is the (component-wise) sum of the covariances of the observed and already mapped cone (Mahalanobis function \cite{enwiki:1027670189})
\begin{align} \label{eq:mahalanobis-weight}
    z_d &= z - \hat{z} \\
    w_m &= \frac{1}{2 \cdot \pi \cdot \sqrt{\det{\Sigma}}} \exp{(-\frac{1}{2} \cdot z_d^T \cdot \Sigma^{-1} \cdot z_d)}
\end{align}

The pose of each particle is calculated by integrating over the velocities ($v_x$, $v_y$ and $r$=yaw rate) and sampling from the pose distribution. Pose weights $w_p$ are calculated independently for x, y and heading by using the Gaussian function\cite{enwiki:1017028878} if GPS is available. The difference is measured between the (integrated) pose and GPS pose and the standard deviations of the GPS are used. This measure is used to avoid (problematic) drifting of the pose.

This results in the total weight for a particle for time step $n+1$ (next observation received) given by (exponents give number of occurrences; $o$ observed cones are matched and $r$ poses are received)
\begin{equation} \label{total-weight}
    w_{t,n+1} = w_{t,n} \cdot w_{no}^i \cdot w_{or}^j \cdot w_{nc}^k \cdot \prod_{l=1}^{o} w_{m,l} \cdot \prod_{q=1}^{r} w_{p,q}
\end{equation}
The Skidpad and Acceleration disciplines are (almost) precisely defined by the competition handbook\cite{FSGHandbook} and are therefore hard-coded in FastSLAM with a particular cone variance. The unknown acceleration track width is determined by assigning each particle a different track width and letting the particle filter sort out the track width and starting pose within two seconds. The pose can be accurately tracked by FastSLAM even if the track is not exactly straight and has slight bends in it. In Skidpad the two circles are defined exactly and are hardcoded in the SLAM map with some variance in all directions (pose is accurate as well).

For Autocross and Trackdrive there are no defined maps. Therefore FastSLAM creates a map in the first round and freezes it for the following laps and goes into localization mode (essentially Monte Carlo Localization, MCL \cite{MCL}) which is sufficient for speeds of up to 30km/h.
%
%
%
%
\subsubsection{Detection of Track Centerline}
As the track layout for Acceleration and Skidpad is well known for these two missions the track centerline is hardcoded and is independent of the detected cones. The Localization algorithm takes care of minor imperfections in the hardcoded cone positions.

Autocross and Trackdrive: To detect the centerline of the track from the cones outputted by localization first of all cones are transformed into track gates. Every cone is assigned one or more other cones and each pair is then saved as a gate formed by those two cones. The position of the gate is the center point of its two cones.
We only consider paths formed by at least 3 gates. To this end, we assign a cost to each path and gate of a path. The cost for a path is the sum of the costs of its gates. With the following approach the cost of a gate depends on the previous 2 gates only which reduces computational resources drastically while maintaining the ability to also detect difficult passages of track.
Starting at the car's starting point every gate within 8m is assigned a cost using a cost function. Cost depends on: distance to the last gate, distance of the two cones forming that gate, required car angle change to pass through the gate, angle at which the car drives through the gate, cone colors.
Color costs are close to zero if cone colors are probable to be correct. Costs are medium if color probabilities are not high (uncertain color). Costs are high if both cones of a gate have the same color and very high if cones have their colors inverted (blue and yellow on the wrong side).

The cheapest 3 gates are used as a new starting point and all gates around those are costed again. This happens a third time to get a tree of depth three and 3 branches at each node.
Now the cheapest path (least sum of cost of nodes) is evaluated and its first node is added to the track centerline. The other two branches are discarded. By costing the gates around the leaves of the now two depth tree, a third depth is added again.

This process continues until the start/finish line is found.

%
%
%
%
\subsubsection{Planning and Control}

Finding the optimal path and the current control outputs is done with Model-Predictive-Control (MPC) techniques. This allows us to reach close to optimal controls that can in the future easily be optimized even more by improving the car model. At each iteration (every 50ms) we solve the following optimization problem for our horizon $H = 40$:
\begin{align*}
    \label{MPC:objective}
    \min_{(\Delta u)_i} & \sum_{k=0}^H q_D \cdot (\Delta D)^2 + q_\phi \cdot \phi^2 - q_{v_x} \cdot v_x - q_{p} \cdot \Delta p + q_s \cdot S_{c,k}^2 + q_d \cdot d \\
    s.t. & x_0 = x(0), \\
        & x_{k+1} = f(x_k, u_k, \Delta u_k), \\
        & d \leq (0.7m)^2 + S_{c,k}, \\
        & B_l \leq G(x_k, u_k, \Delta u_k) \leq B_u \\
     & \!\!\!\!\!\!\!\!\!\!\!\! where \; d = (X_k - X_{cen}(p_k))^2 + (Y_k - Y_{cen}(p_k))^2
\end{align*}
$q_i$ are factors for cost adaption.
We define $u = [\delta, D]^T$ and $\Delta u = [\phi, \Delta D, \Delta p]^T$ with $\delta$ being the steering angle, $\phi$ its derivation, $D$ being the current throttle command between -100 and +100\% and , $\Delta p$ the speed along the track centerline.
Furthermore we define $x = [X, Y, v_x, v_y, \theta, \dot{\theta}, p]^T$ with $X$ and $Y$ being the position of the car in the map, $v_x$ and $v_y$ being the respective velocities, $\theta$ orientation of the car in the map, yawrate $\dot{\theta}$ and progress along the track $p$ in meters ($\dot{p} = \Delta p$).

The vehicle is modelled with a kinematic vehicle model at slow speeds and a dynamic vehicle model at higher speeds. They are blended linearly between 3 and 6 m/s. See \cite{KongPfeifer2015} for models and \cite{AMZFullyAutonomousRacecar} for blending.

$B_l$ and $B_u$ are upper and lower bounds of vehicle dynamics constraints (max velocity, steering angle extrema, \dots) scaled by function $G(\dots)$.
From other modules we know the track centerline, parametrized as points that are equally spaced 4m apart which are interpolated using a cubic spline with spline parameter $p$ with $\Delta p$ being determined during the optimization to always keep the car close to the track.

Negative weighting of $\Delta p$ and $v_x$ encourages progress along the track and minimizes lap times. Penalization of $\Delta u$ smooths the resulting maneuvers and reduces stress on the car.

To improve efficiency MPC is hot started with the last output being the new starting point of the next optimization step.


\section{References}
\foreignlanguage{english}{\printbibliography[heading=none]}

@article{KongPfeifer2015,
    author    = {J. Kong and M. Pfeifer and G. Schildbach and F. Borrelli},
    title     = "{Kinematic and Dynamic Vehicle Models for Autonomous Driving Control Design}",
    journal   = "{IEEE Intelligent Vehicles Symposium (IV)}",
    year      = 2015
}

@other{AMZFullyAutonomousRacecar,
    author    = {J. Kabzan and M. I. Valls and V. J.F. Reijgwart and H. F.C. Hendrikx et al.},
    title     = "{AMZ Driverless: The Full Autonomous Racing System}",
    year      = 2019
}

@other{duan2019centernet,
  title={Centernet: Keypoint triplets for object detection},
  author={Duan, Kaiwen and Bai, Song and Xie, Lingxi and Qi, Honggang and Huang, Qingming and Tian, Qi},
  year={2019}
}

@INPROCEEDINGS{8462829,  author={Valls, Miguel I. and Hendrikx, Hubertus F.C. and Reijgwart, Victor J.F. and Meier, Fabio V. and Sa, Inkyu and Dubé, Renaud and Gawel, Abel and Bürki, Mathias and Siegwart, Roland},  booktitle={2018 IEEE International Conference on Robotics and Automation (ICRA)},   title={Design of an Autonomous Racecar: Perception, State Estimation and System Integration},   year={2018},  volume={},  number={},  pages={2048-2055},  doi={10.1109/ICRA.2018.8462829}}

@article{Zeilinger2017,
    author    = {M. Zeilinger and R. Hauk and M. Bader},
    title     = "{Design of an Autonomous Race Car for the
Formula Student Driverless (FSD)}",
    journal   = "{Proceedings of the OAGM \& ARW Joint Workshop}",
    year      = 2017
}

@webpage{FSGResults2019,
    author = {},
    title = {Results FSG 2019},
    howpublished = {Accessed \url{https://www.formulastudent.de/fsg/results/2019}},
    year = 2019
}

@article{FSGHandbook,
    author = {FSG},
    title = {FSG Handbook 2021},
    url = "https://www.formulastudent.de/fileadmin/user_upload/all/2021/rules/FSG21_Competition_Handbook_v1.0.pdf",
    year = 2021
}

@article{DBLP:journals/corr/abs-1809-10099,
  author    = {Nikhil Bharadwaj Gosala and
               Andreas B{\"{u}}hler and
               Manish Prajapat and
               Claas Ehmke and
               Mehak Gupta and
               Ramya Sivanesan and
               Abel Gawel and
               Mark Pfeiffer and
               Mathias B{\"{u}}rki and
               Inkyu Sa and
               Renaud Dub{\'{e}} and
               Roland Siegwart},
  title     = {Redundant Perception and State Estimation for Reliable Autonomous
               Racing},
  journal   = {CoRR},
  volume    = {abs/1809.10099},
  year      = {2018},
  url       = {http://arxiv.org/abs/1809.10099},
  archivePrefix = {arXiv},
  eprint    = {1809.10099},
  bibsource = {dblp computer science bibliography, https://dblp.org}
}

@article{FastSLAM,
author = {Thrun, Sebastian and Montemerlo, Michael and Koller, Daphne and Wegbreit, Ben and Nieto, Juan and Nebot, Eduardo},
year = {2004},
month = {05},
pages = {},
title = {FastSLAM: An Efficient Solution to the Simultaneous Localization And Mapping Problem with Unknown Data},
volume = {4},
journal = {Journal of Machine Learning Research}
}

@INPROCEEDINGS{MCL,
author={Dellaert, F. and Fox, D. and Burgard, W. and Thrun, S.},
booktitle={Proceedings 1999 IEEE International Conference on Robotics and Automation (Cat. No.99CH36288C)},
title={Monte Carlo localization for mobile robots},
year={1999},
volume={2},
number={},
pages={1322-1328 vol.2},
doi={10.1109/ROBOT.1999.772544}
}

@article{LowVarianceResampling,
author = {Bagnell, Drew and Liu, Tommy},
year = {2012},
title = {Particle Filters: The Good, The Bad, The Ugly},
url = "http://www.cs.cmu.edu/~16831-f14/notes/F11/16831_lecture04_tianyul.pdf"
}

@misc{enwiki:1017028878,
    author = "{Wikipedia contributors}",
    title = "Gaussian function --- {Wikipedia}{,} The Free Encyclopedia",
    year = "2021",
    url = "https://en.wikipedia.org/w/index.php?title=Gaussian_function&oldid=1017028878",
    note = "[Online; accessed 15-May-2021]"
}

@misc{ enwiki:1027670189,
    author = "{Wikipedia contributors}",
    title = "Mahalanobis distance --- {Wikipedia}{,} The Free Encyclopedia",
    year = "2021",
    url = "https://en.wikipedia.org/w/index.php?title=Mahalanobis_distance&oldid=1027670189",
    note = "[Online; accessed 12-June-2021]"
}

@misc{ enwiki:1004599218,
    author = "{Wikipedia contributors}",
    title = "Extended Kalman filter --- {Wikipedia}{,} The Free Encyclopedia",
    year = "2021",
    url = "https://en.wikipedia.org/w/index.php?title=Extended_Kalman_filter&oldid=1004599218",
    note = "[Online; accessed 12-June-2021]"
}

@inproceedings{valls2018design,
  title={Design of an autonomous racecar: Perception, state estimation and system integration},
  author={Valls, Miguel I and Hendrikx, Hubertus FC and Reijgwart, Victor JF and Meier, Fabio V and Sa, Inkyu and Dub{\'e}, Renaud and Gawel, Abel and B{\"u}rki, Mathias and Siegwart, Roland},
  booktitle={2018 IEEE international conference on robotics and automation (ICRA)},
  pages={2048--2055},
  year={2018},
  organization={IEEE}
}

\end{document}